\documentclass{article}

\PassOptionsToPackage{table,xcdraw}{xcolor}
\PassOptionsToPackage{numbers}{natbib}
\usepackage[preprint]{neurips_2025}

\usepackage[utf8]{inputenc} % allow utf-8 input
\usepackage[T1]{fontenc}    % use 8-bit T1 fonts
\usepackage{hyperref}       % hyperlinks
\usepackage{url}            % simple URL typesetting
\usepackage{booktabs}       % professional-quality tables
\usepackage{amsfonts}       % blackboard math symbols
\usepackage{nicefrac}       % compact symbols for 1/2, etc.
\usepackage{microtype}      % microtypography
\usepackage{tikz}
\usetikzlibrary{positioning}
\usepackage{multirow}
\newif\ifshowcomments
\showcommentstrue
\usepackage{adjustbox}
\usepackage{algorithm}
\usepackage{algpseudocode}
\usepackage{amsmath}
\usepackage{amsmath}
\usepackage{booktabs}
\usepackage{enumitem} % Add this in your preamble
\usepackage{wrapfig}
\usepackage{graphicx}
\usepackage{subcaption}

\usepackage{colortbl} % 支持 \rowcolor
\usepackage{array}     % 对齐设置
\usepackage{makecell}  % 表格内多行或加粗支持
\usepackage{caption}   % 控制 caption 样式（可选）
\usepackage{tabularx}
\newcolumntype{C}{>{\centering\arraybackslash}X}  % 自动均分、居中
\ifshowcomments

\title{Silence is Not Consensus: \\ Disrupting Agreement Bias in Multi‑Agent LLMs via Catfish Agent for Clinical Decision Making}

\author{
Yihan Wang$^{1,}$\thanks{These authors contributed equally.}~, 
Qiao Yan$^{1,}$\footnotemark[1]~, 
Zhenghao Xing$^{1,}$\footnotemark[1]~~\thanks{Project lead.}~, 
Lihao Liu$^{2,}$\thanks{Co-corresponding author: \texttt{lihaoliu@amazon.com}}~, 
Junjun He$^{3}$, \\
\textbf{Chi-Wing Fu}$^{1}$, 
\textbf{Xiaowei Hu}$^{3,}$\thanks{Primary corresponding author: \texttt{huxiaowei@pjlab.org.cn}}~, 
\textbf{and} \textbf{Pheng-Ann Heng}$^{1}$ \\[1ex]
$^{1}$The Chinese University of Hong Kong \\
$^{2}$Amazon
$^{3}$Shanghai Artificial Intelligence Laboratory
}

\begin{document}

\maketitle

\setcounter{footnote}{0}
\begin{abstract}
Large language models (LLMs) have demonstrated strong potential in clinical question answering, with recent multi-agent frameworks further improving diagnostic accuracy via collaborative reasoning. 
However, we identify a recurring issue of  \textbf{Silent Agreement}, where agents prematurely converge on diagnoses without sufficient critical analysis, particularly in complex or ambiguous cases.
We present a new concept called \textbf{Catfish Agent}, a role-specialized LLM designed to inject structured dissent and counter silent agreement. 
Inspired by the ``catfish effect'' in organizational psychology, the Catfish Agent is designed to challenge emerging consensus to stimulate deeper reasoning. 
We formulate two mechanisms to encourage effective and context-aware interventions: (i) a complexity-aware intervention that modulates agent engagement based on case difficulty, and (ii) a tone-calibrated intervention articulated to balance critique and collaboration.
Evaluations on nine medical Q\&A and three medical VQA benchmarks show that our approach consistently outperforms both single- and multi-agent LLMs frameworks, including leading commercial models such as GPT-4o and DeepSeek-R1.
\end{abstract}    
\section{Introduction}
\label{sec:intro}

\begin{quote}
\emph{“Without contraries is no progression.”} — \textsc{William Blake}
\end{quote}

Progress often emerges not from agreement but from conflict, when ideas clash and debate arises, before a better solution is derived. This insight resonates deeply in collaborative reasoning.

Large Language Models (LLMs) have demonstrated strong potential in medical diagnosis by leveraging extensive clinical knowledge~\cite{singhal2023large}. 
To enhance diagnostic robustness, recent work has proposed LLM-based multi-agent frameworks, where multiple specialized agents interact to simulate medical teamwork~\cite{tang2023medagents,kim2024mdagents,wang2024beyond,chen2025enhancing}. By fostering diverse reasoning paths and encouraging dissent, these frameworks aim to improve decision quality, particularly in complex cases.

\begin{figure}[tp]
\centering
\includegraphics[width=0.999\linewidth]{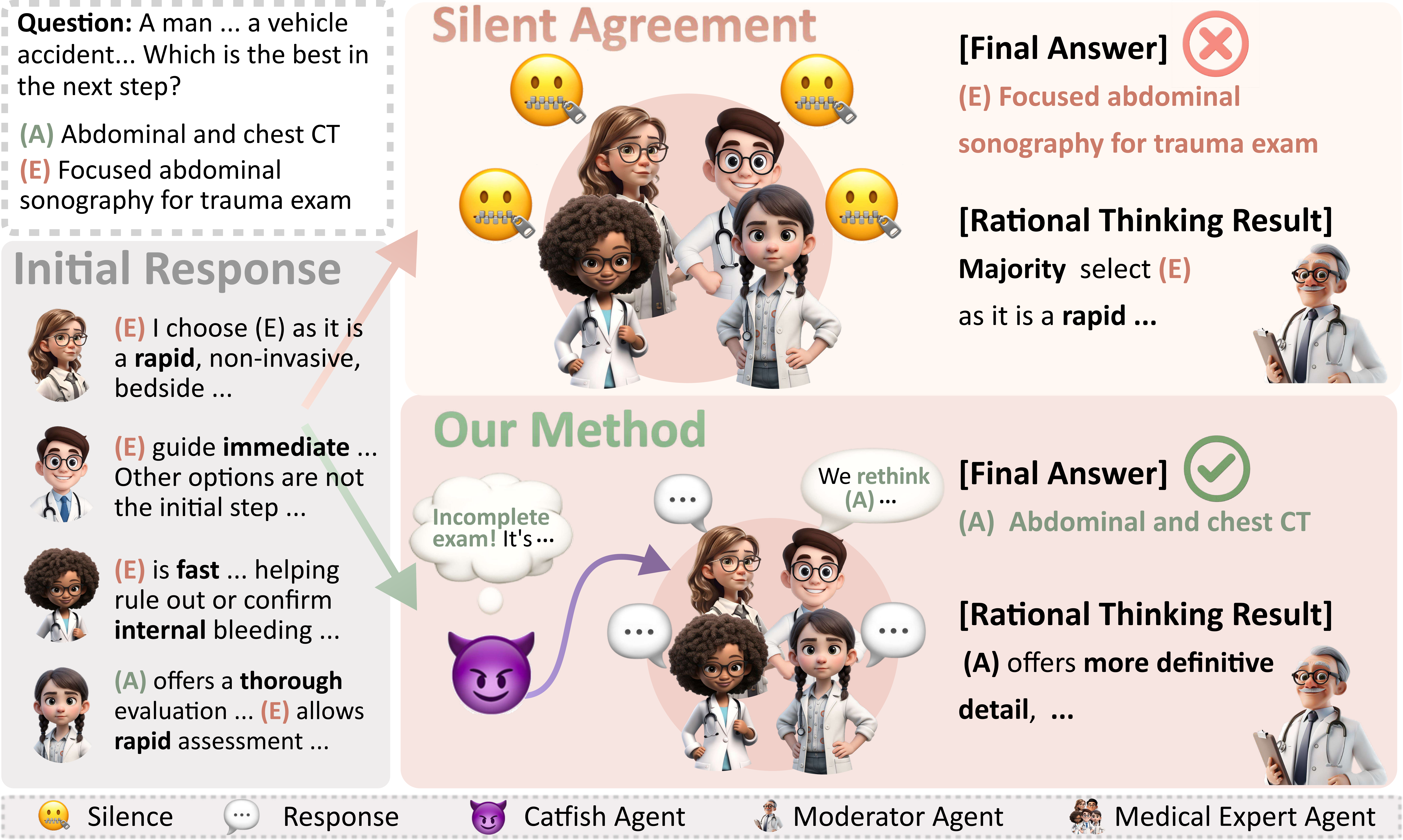}
    \caption{An example clinical misdiagnosis case resulted from \textbf{Silent Agreement}. Although the agents initially select different options, they remained silent in subsequent discussion, resulting in the misdiagnosis. Our method actively disrupts such silent agreement with the designated catfish agent in multi-agent collaborative reasoning and successfully produces the correct outcome.}
\label{fig:silent-example}
\vspace{-5mm}
\end{figure}

However, achieving effective collaboration among LLM agents for clinical decision making remains a significant challenge. 
In practice, we observe a phenomenon we call \textbf{Silent Agreement}, \emph{where a group of medical agents converge prematurely on the same diagnosis, without debate, evaluation, or exploration of alternatives.}
Figure~\ref{fig:silent-example} shows an example clinical misdiagnosis caused by Silent Agreement. Although the agents initially propose different options, no further perspectives are offered, and all agents remain silent in the discussion, ultimately leading to an incorrect diagnosis.

Silent Agreement mirrors a classic failure mode of human groups, often called ``groupthink'', where individuals suppress dissent and converge on a superficial consensus, often leading to sub-optimal or even dangerous decisions \cite{janis1982groupthink}. 
In contrast, social science research shows that constructive disagreement can enhance group performance, especially in high-stakes domains like medicine, by surfacing overlooked evidence and reducing errors~\cite{nemeth1995dissent,mercier2017enigma}. 
Structured dissent and open debate have also been linked to more robust scientific outcomes and deeper reasoning in collaborative settings~\cite{sarewitz2011voice}.
Motivated by these findings, we investigate how dissent can mitigate premature consensus in multi-agent clinical decision making. We identify Silent Agreement as a critical bottleneck, highlighting the need for deeper reasoning by promoting regulated, constructive disagreement.

In this paper, we develop a new concept, namely \textbf{Catfish Agent}, which is designed to \emph{actively disrupt silent agreement in multi-agent collaborative reasoning for clinical decision making.} 
Inspired by the ``catfish effect''\footnote{\url{https://en.wikipedia.org/wiki/Catfish_effect}: The practice of placing a catfish in a tank of sardines to keep them alive. Without stimulation, sardines often become sluggish and suffocate to death due to lack of oxygen; however, the presence of a catfish keeps them constantly moving and thus alive.} and the ``devil’s advocate'' strategy in organizational psychology research~\cite{macdougall1997devil, nemeth2001devil, akhmad2021closed}, we propose to organize multi-agent reasoning as a multi-round, multi-role process. 

Integrating dissent into medical agent groups poses two key challenges: (i) the level of required autonomy varies with case complexity, and (ii) overly assertive dissent can derail discussion or obscure key evidence.
To address these issues, we formulate two core mechanisms in Catfish Agent: 
\emph{(i) Complexity-aware intervention, i.e.,} the agent adapts its engagement based on task difficulty, increasing autonomy in more complex cases to encourage deeper reasoning, and 
\emph{(ii) Tone-calibrated intervention,} in which the strength and tone of dissent vary with the level of agent agreement, avoiding both passivity and excessive disruption.
These novel mechanisms encourage the Catfish Agent to ``break the silence,'' while preserving productive collaboration.

Figure~\ref{fig:silent-example} shows an example case, where the Catfish Agent disrupts premature consensus by critically challenging the expert assumptions. 
This intervention prompts a revision of initial reasoning and enables the framework to synthesize a more reliable diagnosis.
We evaluate our method on nine medical question-answering (Q\&A)~\cite{jin2021disease,jin2019pubmedqa,pal2022medmcqa,wang2024mmlu,kim2024medexqa,zuo2025medxpertqa,hendrycks2020measuring,chen2024benchmarking} and three medical visual question-answering (VQA) benchmarks~\cite{zuo2025medxpertqa,zhang2023pmc,he2020pathvqa}, comparing it with both single-agent LLMs (\textit{e.g.}, GPT-4o~\cite{achiam2023gpt}, DeepSeek-R1~\cite{guo2025deepseek}, HuatuoGPT-o1~\cite{chen2024huatuogpt}) and multi-agent medical frameworks (\textit{e.g.}, MedAgent~\cite{tang2023medagents}, MDAgent~\cite{kim2024mdagents}). 
Experimental results show that our method achieves a 12.73-point improvement on average, corresponding to a 39.2\% relative gain over the best prior model, DeepSeek-R1, on the Q\&A benchmarks, and a 5.33-point improvement on average, representing a 12.7\% relative gain over the best prior method, MDAgent, on the VQA benchmarks.
\emph{We will release our code, experimental results, and logs.} Our contributions are threefold:
\begin{itemize}[leftmargin=*]
    \item We identify and formally define the \textbf{Silent Agreement} problem in LLM-based multi-agent frameworks for clinical decision making.
    
    \item We present the new concept \textbf{Catfish Agent}, the first to inject structured dissent into medical multi-agent systems, using the proposed \textit{complexity-aware} and \textit{tone-calibrated} interventions to break Silent Agreement and enhance collaborative clinical reasoning.
    
    \item We conduct extensive experiments on nine medical Q\&A and three medical VQA benchmarks, demonstrating that our method largely outperforms state-of-the-art single- and multi-agent models.
\end{itemize}

\section{Related Works}
\label{sec:relatedwork}

\textbf{Multi-Agent LLM for Medical Decision Making (MDM).} 
Recent studies have applied multi-agent LLM frameworks to collaborative tasks in planning, coding, and healthcare~\cite{zhang2024towards,wu2023autogen,kim2024mdagents}, typically assigning complementary roles to agents to support multi-turn coordination. AutoGen~\cite{wu2023autogen} formalizes inter-agent communication for iterative reasoning, while MDAgents~\cite{kim2024mdagents} models medical teams with role-specialized agents for diagnostic support. However, most approaches emphasize cooperation over critique, assuming alignment leads to better decisions. In practice, we identify a critical failure mode called \textit{Silent Agreement}, where agents prematurely converge on diagnoses without considering alternative hypotheses or resolving evidence conflicts. 
Recent works explore multi-agent debate~\cite{wang2024rethinking,liubreaking,kim2024debate,wang2024devil}, but often overlook silent agreement. 

\textbf{Large Language Models for Medical Reasoning.} LLMs have demonstrated growing capabilities in clinical reasoning, question answering, and medical summarization tasks \cite{nori2023capabilities,luo2024biomedgpt,wang2025baichuan}. 
Models like \cite{achiam2023gpt, guo2025deepseek,chen2024huatuogpt,wang2025baichuan,openai2024gpt4omini,openai2024openaio1mini,liu2024deepseek,openai2025openaio3mini,qwen2025qwq32b,grattafiori2024llama,claude2024claude35} have achieved strong performance on benchmarks like MedQA \cite{jin2021disease} and PubMedQA \cite{jin2019pubmedqa}. 
Recent work has begun exploring interaction-based improvements, including CoT prompting and collaborative diagnosis \cite{liu2024medcot,kim2024mdagents}. 
Nevertheless, current methods rarely address the group-level dynamics of agreement or disagreement.
To the best of our knowledge, this is the first work to identify and mitigate the silent agreement bias in LLM-based multi-agent frameworks for medical decision making through a new structured role-based intervention, \textit{i.e.}, Catfish Agent.

\section{The Silent Agreement Problem}
\label{sec:problem}

We start this research work by carefully studying the prevalence and impact of Silent Agreement, a critical failure mode in multi-agent medical LLM frameworks, where agents converge on an answer, often incorrect, without sufficient deliberation or justification. This behavior undermines the intended collaborative nature of multi-agent reasoning and introduces risks in medical decision making.

To assess this issue, we analyze the \textit{hard} set from MedAgentBench~\cite{tang2025medagentsbench}, focusing on two widely-used benchmarks: MedQA~\cite{jin2021disease} and PubMedQA~\cite{jin2019pubmedqa}. We evaluate two prominent multi-agent frameworks, MedAgents~\cite{tang2023medagents} and MDAgents~\cite{kim2024mdagents}, along with our proposed method. A \textit{silent agreement failure} is defined as a diagnostic error, where agents produce a final answer without meaningful discussion, critique, or verification.

As shown in Table~\ref{tab:silent_error_comparison}, MedAgents and MDAgents exhibit high silent rates, over 61.0\% on both datasets, indicating frequent non-response or unjustified consensus. More critically, a large portion of their diagnostic failures are attributable to silent agreement: for MedAgents, 61.9\% of failures on MedQA and 90.7\% on PubMedQA; for MDAgents, 68.1\% and 64.0\%, respectively. These patterns confirm that silent agreement is not a rare anomaly but a dominant failure type in existing methods.
Chi-squared tests~\cite{pearson1900x} further confirm that silent agreement significantly impacts diagnostic accuracy in both frameworks: MDAgents ($\chi^2(1)=5.345$, $p=0.0208$) and MedAgents ($\chi^2(1)=5.896$, $p=0.0152$), revealing a strong association between silent agreement and diagnostic failures.\footnote{Statistically significant at $p<0.05$}

In contrast, our method achieves a significantly lower silent rate: 17.0\% on MedQA and 11.0\% on PubMedQA. Moreover, among the failures, our method makes only 18.0\% and 14.3\% involve silent agreement, substantially lower than those of MedAgents and MDAgents. This result indicates that our framework not only reduces unjustified silence but also encourages agents to engage in meaningful deliberation. Importantly, shifting away from silent behavior aligns with improved diagnostic accuracy, as our method outperforms existing multi-agent frameworks; see Table~\ref{tab:multi-agent} for more details. The underlying mechanism is detailed in the next section.

\begin{table*}[tp!]
\centering
\caption{Silent behavior analysis across MedQA and PubMedQA. ``Silent Rate'' denotes the proportion of questions, where agents arrive at a final answer with silent agreement. ``Failure Attribution Rate'' refers to the proportion of diagnostic failures that result from silent agreement. Our method achieves both the lowest silent rate and the lowest attribution to silent agreement failures.}
\label{tab:silent_error_comparison}
\vspace{-1mm}
\begin{subtable}[t]{0.495\textwidth}
    \centering
    \caption{MedQA Dataset.}
    \vspace{-1mm}
    \label{tab:medqa_silent_error}
    \begin{adjustbox}{width=0.9\linewidth}
    \begin{tabular}{lcc}
        \toprule
        \textbf{Method} & \textbf{Silent Rate} $\downarrow$ & \textbf{Failure Attr. Rate} $\downarrow$ \\
        \midrule 
        % \midrule
        MedAgents & 64.0\% & 61.9\% \\
        MDAgents & 61.0\% & 68.1\% \\
        \textbf{Ours}      & \textbf{17.0\%} & \textbf{18.0\%} \\
        \bottomrule
    \end{tabular}
    \end{adjustbox}
    \vspace{-2mm}
\end{subtable}
\hfill
\begin{subtable}[t]{0.499\textwidth}
    \centering
    \caption{PubMedQA Dataset.}
    \vspace{-1mm}
    \label{tab:pubmedqa_silent_error}
    \begin{adjustbox}{width=0.9\linewidth}
    % \begin{tabular}{l||cc}
    \begin{tabular}{lcc}
        \toprule
        \textbf{Method} & \textbf{Silent Rate} $\downarrow$ & \textbf{Failure Attr. Rate} $\downarrow$ \\
        \midrule 
        % \midrule
        MedAgents  & 89.0\% & 90.7\% \\
        MDAgents & 61.0\% & 64.0\% \\
        \textbf{Ours}      & \textbf{11.0\%} & \textbf{14.3\%} \\
        \bottomrule
    \end{tabular}
    \end{adjustbox}
\vspace{-5mm}
\end{subtable}
\end{table*}

\section{Catfish Agent: Breaking Silent Agreement in LLM Teams}
\label{sec:method}

To address the Silent Agreement problem in LLM-based multi-agent clinical reasoning, we draw inspiration from organizational psychology, where structured disagreement has been shown to enhance epistemic vigilance and decision accuracy in human teams.
Translating this principle to LLM-based teams introduces two key challenges: (i) \emph{the necessary level of dissent varies with case complexity}, and (ii) \emph{overly aggressive disagreement can derail discussion or obscure key evidence.}

To address these challenges, we design the Catfish Agent with two core mechanisms:  
(i) a \emph{complexity-aware intervention strategy} that adapts the agent’s behavior to the difficulty of the clinical case, and  
(ii) a \emph{tone-calibrated intervention mechanism} that adjusts the rhetorical strength of dissent based on the level of group agreement.
The complexity-aware intervention strategy adaptively controls the Catfish Agent’s degree of autonomy based on task difficulty (basic, intermediate, advanced), as assessed by the Moderator. It governs \emph{when} and \emph{how much} the agent should intervene in potential groupthink. In contrast, the tone-calibrated intervention mechanism determines \emph{how} the dissent is expressed, ensuring interventions are context-sensitive and constructively framed.
Figure~\ref{fig:pipeline} illustrates the overall workflow of our framework involving the Catfish Agent, while Sections~\ref{sec:disturbance-strategy} and~\ref{sec:intervention-tone} present the details in the two core mechanisms.

\subsection{Catfish in the Tiers: Stratifying Intervention by Complexity}
\label{sec:disturbance-strategy}

Clinical tasks vary in complexity, with simple cases yielding quick consensus and complex cases requiring deeper reasoning. 
Therefore, this \emph{complexity-aware intervention strategy} is proposed to dynamically adjusts the Catfish Agent’s behavior based on case difficulty. In complex or ambiguous scenarios, the Catfish Agent is granted a stronger sense of independent judgment. Conversely, in simpler cases, its interventions are more limited and guided. The following describes how the Catfish Agent’s behavior is progressively liberated according to different levels of case complexity.

\vspace{-2.5mm}
\paragraph{Basic cases.} 
For low-complexity clinical questions, the Moderator independently formulates an initial diagnosis $D$.
The Catfish Agent then performs a lightweight critique, reviewing the reasoning behind $D$ to identify any overlooked differentials or incomplete justification. If meaningful issues are detected, it generates a comment for the Moderator’s reference when making the final diagnosis $F$.

\vspace{-2.5mm}
\paragraph{Intermediate cases.}
In these cases, the Moderator first recruits a group of expert agents $A$. 
Specifically, the Catfish Agent $C$ is recruited alongside other agents, who is restricted to a predefined area of expertise due to its assigned role as determined by the Moderator. As shown in Figure~\ref{fig:silent-example}, during the group debate, the Catfish Agent is responsible for monitoring group dynamics and introducing dissent when appropriate.

The reasoning process unfolds over multiple rounds, each consisting of several turns.
First, before the initial round begins, all recruited agents, excluding $C$, independently generate initial diagnoses $D$, which are sequentially shared for peer review.
Second, in each round $i$, the process proceeds through a series of turns $t$. In each turn, agents sequentially evaluate the latest collective responses $R^{i,t-1}$ and contribute updated outputs $R^{i,t}$ based on their domain expertise. The first turn of the first round is grounded in $D$.
Third, $C$ monitors turn-level responses for logical inconsistencies, missed differentials, or weak justifications, aiming to detect emerging Silent Agreement. Upon identifying such issues, it injects domain-specific perturbations as assigned by the Moderator.
Fourth, the rhetorical strength of these interventions (\textit{e.g.}, mild, moderate, strong) is modulated based on the perceived degree of group convergence. Agents targeted by these challenges may revise their responses if they find the intervention sufficiently compelling.
Last, at the end of each round, a Summary Agent compiles a structured report $R^i$ that aggregates the group’s updated reasoning.

The reasoning process terminates under one of two conditions: (i) all agents reach consensus or fall into Silent Agreement, and two consecutive interventions by the Catfish Agent fail to provoke meaningful divergence; or (ii) the discussion reaches a predefined limit of $n$ rounds, with each round allowing up to $t$ interaction turns.
Last, the Moderator reviews the interaction log and optionally consults intermediate summaries. 
The Catfish Agent may intervene if superficial consensus or diagnostic gaps emerge. 
The final decision $F$ is made by the Moderator, integrating cross-round insights and catfish feedback with critical awareness.

\begin{figure}[tp!] % 'ht' means here or top
    \centering % Center the image
    \includegraphics[width=0.999\textwidth]{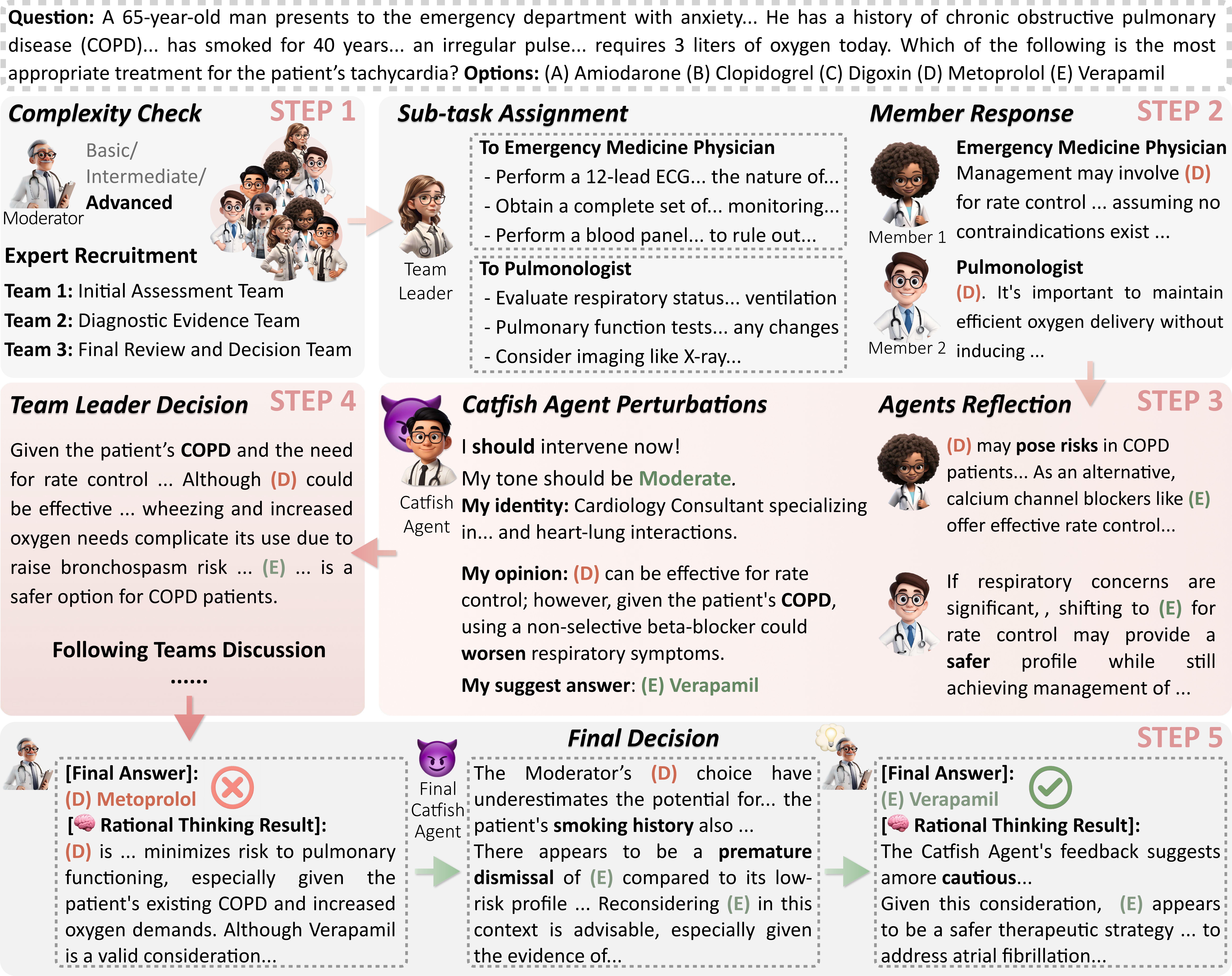} 
    \caption{Overview of the reasoning process for an advanced case. 
   (i) the system routes the clinical question through a complexity-aware Moderator, which classifies it as advanced and activates three expert teams, each consisting of a leader and two members;
(ii) within each team, the leader assigns specific subtasks, and members respond independently based on their expertise;
(iii) a Catfish Agent monitors the discussion and selectively intervenes by critiquing flawed assumptions or incomplete reasoning. All team members are required to respond to these challenges;
(iv) after internal discussion, each team leader finalizes the team's answer and forwards it to the next team for iterative refinement; and
(v) once all teams have contributed, the Moderator synthesizes the collective reasoning and, if needed, introduces an additional Catfish Agent for final diagnosis.
}
\label{fig:pipeline} % Label for referencing the image
\vspace{-5mm}
\end{figure}

\vspace{-3.5mm}
\paragraph{Advanced cases.} 
For high-complexity diagnostic tasks, the Moderator initiates a hierarchical team-of-teams structure.  
As illustrated in Figure~\ref{fig:pipeline}, multiple teams $t_j = \{a_1, a_2, \dots, a_m\}$ are recruited, each composed of domain-specific agents and led by a team leader. Within each team, the leader assigns sub-tasks to members, who collaborate to generate a structured report. Teams then take turns presenting their findings, allowing for cross-team comparison and integrative reasoning.

In the highest-complexity scenarios, the Catfish Agent $C^\ast$ evolves into a free-roaming entity capable of independently initiating dissent with self-determined roles and perspectives.
First, upon detecting Silent Agreement or conversational bias, $C^\ast$ adopts an appropriate medical persona (\textit{e.g.}, a Senior Gastrointestinal Oncologist with 20+ years of experience in colorectal malignancies) and issues context-sensitive challenges or counterarguments.
Second, these interventions are integrated into the team’s workflow, resulting in updated reasoning traces.
Third, the team’s structured report is then passed sequentially to the next team $t_{j+1}$, enabling cumulative refinement across teams. Unlike the fixed-role Catfish $C$ in intermediate settings, $C^\ast$ dynamically traverses teams, contributing from a broader, cross-domain perspective rather than being constrained to a predefined specialty.
Last, once all $m$ teams have completed their contributions, the Moderator synthesizes the aggregated insights and, if necessary, introduces a final Catfish Agent to conduct cross-team critique before producing the final diagnosis $F$. 
This process ensures both in-depth intra-team deliberation and cross-team integration before final decision making.

To sum up, benefiting from our \textit{complexity-aware} intervention strategy, the Catfish Agent exhibits a progressively independent mindset and structural influence across basic, intermediate, and advanced cases, aligning with the increasing complexity of diagnostic scenarios.

\subsection{Catfish in the Tone: Scaling Dissent by Consensus Strength}\label{sec:intervention-tone}

In collaborative diagnostic reasoning, the challenge is not only deciding \emph{when} and \emph{how much} to introduce dissent, but also \emph{how} to express it, specifically, determining the appropriate degree of rhetorical intensity to ensure disagreement is constructive rather than disruptive.
Early convergence among agents may indicate either well-founded consensus or premature closure. The core challenge lies in calibrating dissent: \emph{if too weak, it may be ignored; if too strong, it risks disrupting constructive reasoning or derailing the diagnostic process}.

This tone-calibrated intervention mechanism is proposed to address this underexplored problem. It can allows the Catfish Agent dynamically adjusts the rhetorical intensity in response to the strength of inter-agent consensus. 
This mechanism enables context-sensitive dissent that preserves epistemic rigor without compromising team coherence.
We introduce different tones of intervention as follows:

\vspace{-2.5mm}
\paragraph{Mild interventions.} This type of intervention targets scenarios in which agents begin to converge prematurely, relying on minimal or shallow evidence.
In such cases, the Catfish Agent initiates reflective, non-confrontational prompts 
to gently encourage broader consideration. 
These interventions aim to stimulate metacognitive reflection without disrupting group cohesion. %\yihan{updated}

\vspace{-2.5mm}
\paragraph{Intermediate interventions.} This level of intervention is intended for scenarios in which conclusions are formed without robust supporting evidence.
The Catfish Agent probes with targeted questions, 
applying constructive pressure to surface hidden assumptions. This strategy draws inspiration from Socratic inquiry and diagnostic timeout protocols \cite{ho2023thinking,graber2012cognitive} in the field of human clinical decision making.

\vspace{-2.5mm}
\paragraph{Strong interventions.} This level of intervention targets cases of rapid, uncritical agreement among agents. In such cases, the Catfish Agent delivers assertive challenges, explicitly questioning the group’s reasoning to counteract conformity bias. This mode aligns with cognitive conflict theory and the devil’s advocate paradigm.

Overall, this mechanism scales dissent according to consensus dynamics, ensuring interventions are epistemically productive and collaboration-preserving, while avoiding unnecessary disruption in simple scenarios and intensifying epistemic friction in high-stakes, complex cases.

\section{Experimental Results}
\label{sec:exp}

\begin{table}[tp]
\caption{Comparison results on medical Q\&A datasets. Accuracy (\%) is shown per task, with the \textit{hard} set categorized according to MedAgentsBench. Models are grouped into \textbf{general-domain} (upper block) and \textbf{medical-domain} (middle block) categories, followed by our method. \textbf{Best} results are in bold; \underline{second-best} are underlined. Our method is based on the o3-mini model.}
\label{tab:fina_hard}
\renewcommand{\arraystretch}{1.2}
\fontsize{10pt}{11pt}\selectfont
\begin{adjustbox}{width=\textwidth}
\begin{tabular}{p{2.8cm}
% ||
>{\centering\arraybackslash}p{1.3cm} >{\centering\arraybackslash}p{1.3cm} >{\centering\arraybackslash}p{1.3cm} >{\centering\arraybackslash}p{1.3cm} >{\centering\arraybackslash}p{1.3cm} >{\centering\arraybackslash}p{1.3cm} >{\centering\arraybackslash}p{1.3cm} >{\centering\arraybackslash}p{1.3cm} >{\centering\arraybackslash}p{1.3cm}}
\toprule
\textbf{Method} & 
\makecell{\textbf{Med}\\\textbf{QA}} & \makecell{\textbf{PubMed}\\\textbf{QA}}  & \makecell{\textbf{Med}\\\textbf{MCQA}} &
\makecell{\textbf{Med}\\\textbf{Bullets}} & 
\textbf{MMLU} & 
\makecell{\textbf{MMLU}\\\textbf{-Pro}} &
\makecell{\textbf{MedEx}\\\textbf{QA}} &
\makecell{\textbf{MedX}\\\textbf{pert-R}} &
\makecell{\textbf{MedX}\\\textbf{pert-U}} \\
\midrule
% \midrule
GPT-4o-mini~\cite{openai2024gpt4omini} & 22.0 & 10.0 & 17.0 & 10.1 & 12.3 & 11.0 & 4.0 & 6.0 & 5.0 \\
GPT-4o~\cite{openai2024hello} & 32.0 & 9.0 & 25.0 & 19.1 & 24.7 & 21.0 & 18.0 & 7.0 & 6.0 \\
DeepSeek-V3~\cite{liu2024deepseek} & 16.0 & 12.0 & 19.0 & 13.5 & 15.1 & 12.0 & 7.0 & 6.0 & 9.0 \\
o1-mini~\cite{openai2024openaio1mini} & 49.0 & 11.0 & 21.0 & 38.2 & 31.5 & 19.0 & 15.0 & \underline{29.0} & 14.0 \\
o3-mini~\cite{openai2025openaio3mini} & \underline{53.0} & \underline{16.0} & 24.0 & \underline{50.6} & 35.6 & 15.0 & 18.0 & 25.0 & 15.0 \\
QwQ-32B~\cite{qwen2025qwq32b} & 29.0 & \underline{16.0} & 24.0 & 12.4 & 19.2 & 28.0 & 10.0 & 9.0 & 6.0 \\
DeepSeek-R1~\cite{guo2025deepseek} & 47.0 & 13.0 & \underline{31.0} & 43.8 & \underline{43.8} & \underline{37.0} & \underline{26.0} & 25.0 & \underline{26.0} \\
Llama-3.3-70B~\cite{grattafiori2024llama} & 14.0 & 13.0 & 20.0 & 16.9 & 12.3 & 10.0 & 7.0 & 9.0 & 9.0 \\
Claude-3.5-S~\cite{claude2024claude35} & 18.0 & 10.0 & 10.0 & 9.0 & 16.4 & 14.0 & 13.0 & 9.0 & 11.0 \\
Claude-3.5-H~\cite{claude2024claude35} & 13.0 & 12.0 & 23.0 & 10.1 & 11.0 & 12.0 & 13.0 & 8.0 & 6.0 \\
\midrule
% \midrule
HuatuoGPT-o1~\cite{chen2024huatuogpt}  & 28.0 & 15.0 & \underline{31.0} & 10.1 & 17.8 & 28.0 & 8.0 & 7.0 & 4.0 \\
Baichuan-M1~\cite{wang2025baichuan} & 20.0 & 13.0 & 22.0 & 14.6 & 15.1 & 21.0 & 9.0 & 6.0 & 5.0 \\
\midrule
% \midrule
\rowcolor[HTML]{EFEFEF}
\textbf{Ours} & \textbf{62.0} & \textbf{34.0} & \textbf{45.0} & \textbf{66.3} & \textbf{47.9} & \textbf{48.0} & \textbf{33.0} & \textbf{37.0} & \textbf{34.0} \\
\textit{Improvements} & \textit{+9.0} & \textit{+18.0} & \textit{+14.0} & \textit{+15.7} & \textit{+4.1} & \textit{+11.0} & \textit{+7.0} & \textit{+8.0} & \textit{+8.0} \\
\bottomrule
\end{tabular}
\end{adjustbox}
\vspace{-2mm}
\end{table}

In this section, we evaluate the effectiveness of our proposed Catfish Agent Framework on MedAgentsBench \cite{tang2025medagentsbench}, a benchmark designed to assess complex medical reasoning. 
MedAgentsBench is built from eight diverse medical Q\&A datasets, including MedQA \cite{jin2021disease}, PubMedQA \cite{jin2019pubmedqa}, MedMCQA \cite{pal2022medmcqa}, MedBullets \cite{chen2024benchmarking}, MedExQA \cite{kim2024medexqa}, and MedXpertQA \cite{zuo2025medxpertqa}. Note that MedXpertQA consists of MedXpert-U and MedXpert-R, with each subset focusing on understanding and reasoning. It also integrates six medical tasks from MMLU \cite{hendrycks2020measuring} and MMLU-Pro \cite{wang2024mmlu}. Based on performance and reasoning depth, challenging ``hard'' subsets are selected. 
For a fair comparison, we follow the standardized evaluation protocol and use the officially results reported by MedAgentsBench \cite{tang2025medagentsbench}.

\vspace{-2.5mm}
\paragraph{Implementation details.}
All experiments are conducted via the OpenAI API\footnote{\url{https://platform.openai.com}} in a strict zero-shot setting, without any fine-tuning or gradient updates.
Each agent, the Moderator, Catfish Agent, and domain-specific Experts, is instantiated through separate API calls, with roles defined by structured prompts incorporating system instructions and dialogue history.
Default API parameters (\texttt{temperature}, \texttt{top\_p}) are employed, with no explicit constraint on \texttt{max\_tokens}.

\subsection{Comparison with General and Medical Large Models}

We compare our method using o3-mini~\cite{openai2025openaio3mini} as the base model for each agent with the state-of-the-art general large models, including GPT-4o-mini~\cite{openai2024gpt4omini}, GPT-4o~\cite{openai2024hello}, DeepSeek-V3~\cite{liu2024deepseek}, o1-mini~\cite{openai2024openaio1mini}, o3-mini~\cite{openai2025openaio3mini}, QwQ-32B~\cite{qwen2025qwq32b}, DeepSeek-R1~\cite{guo2025deepseek}, Llama-3.3-70B~\cite{grattafiori2024llama}, Claude-3.5-S~\cite{claude2024claude35}, and Claude-3.5-H~\cite{claude2024claude35}, as well as specific medical models, including HuatuoGPT-o1~\cite{chen2024huatuogpt} and Baichuan-M1~\cite{wang2025baichuan}.
\emph{Notably, we are the first to integrate a reasoning model, namely o3-mini, into a multi-agent framework for medical decision making.}

Table~\ref{tab:fina_hard} presents the performance of our approach on MedAgentsBench, in comparison with a broad range of general-purpose and medical-specialized large language models, focusing on the ``hard'' subsets requiring demand deeper reasoning.
Our method consistently achieves state-of-the-art performance across all benchmarks, surpassing the second-best model by a substantial margin, \emph{yielding an average 12.7-point absolute gain, corresponding to a 39.2\% relative improvement\footnote{The overall average accuracy improvement is computed by first averaging accuracy across all tasks and then calculating the relative gain over DeepSeek-R1, resulting in a 39.2\% improvement.} over the best prior model, DeepSeek-R1}.

Additionally, we have the following observations.
(i) \textbf{Our method is the first to integrate CoT-style reasoning into a structured multi-agent framework that supports multi-turn deliberation under complex clinical conditions.} 
By embedding CoT reasoning into each agent’s decision process and introducing structured dissent via a Catfish Agent, our system not only improves diagnostic accuracy but also offers a novel paradigm for modeling disagreement, iterative reasoning, and collaboration, which are the key characteristics of expert clinical teams.
(ii) \textbf{Reasoning-based LLMs substantially outperform standard LLMs across all medical benchmarks.} 
For example, \texttt{o3-mini} and \texttt{DeepSeek-R1} achieve significantly higher accuracy than their non-reasoning counterparts, indicating that explicit intermediate reasoning steps, such as Chain-of-Thought (CoT), are highly effective in complex medical tasks.
(iii) \textbf{General-purpose reasoning models consistently outperform domain-specialized medical LLMs.} 
Despite lacking medical-specific pretraining, models like \texttt{o3-mini} and \texttt{DeepSeek-R1} surpass medical-tuned models such as \texttt{HuatuoGPT-o1} and \texttt{Baichuan-M1} across nearly all datasets. 
This suggests that broad reasoning capabilities provide greater benefits than narrow domain knowledge, especially in high-level diagnostic tasks such as MedXpertQA. 

\begin{table}[tp]
  \caption{Comparison results on medical Q\&A datasets. All tasks are evaluated on the \textit{hard} set, with accuracy reported in percentage (\%). Two base models are used: GPT-4o-mini and GPT-4o. \textbf{Best results} are in bold; \underline{second-best} are underlined. Methods are grouped into four categories (Baseline-Prompting, Advanced-Prompting, Search-Agent, Multi-Agent).
  }
  \label{tab:multi-agent}
  \centering
  \small
  \renewcommand{\arraystretch}{1.3}
  \setlength{\tabcolsep}{3pt}
  \resizebox{\textwidth}{!}{%
  \begin{tabular}{
        @{\hspace{\tabcolsep}}l@{\hspace{\tabcolsep}}
        % || 
        *{9}{cc}
      @{}}
      \toprule
      \multirow{2}{*}{\textbf{Method}}
        & \multicolumn{2}{c}{\makecell{\textbf{Med}\\\textbf{QA}}}
        & \multicolumn{2}{c}{\makecell{\textbf{PubMed}\\\textbf{QA}}}
        & \multicolumn{2}{c}{\makecell{\textbf{Med}\\\textbf{MCQA}}}
        & \multicolumn{2}{c}{\makecell{\textbf{Med}\\\textbf{Bullets}}}
        & \multicolumn{2}{c}{\textbf{MMLU}}
        & \multicolumn{2}{c}{\makecell{\textbf{MMLU}\\\textbf{-Pro}}}
        & \multicolumn{2}{c}{\makecell{\textbf{MedEx}\\\textbf{QA}}}
        & \multicolumn{2}{c}{\makecell{\textbf{Med}\\\textbf{Xpert-R}}}
        & \multicolumn{2}{c}{\makecell{\textbf{Med}\\\textbf{Xpert-U}}} \\
      \cmidrule(lr){2-3} \cmidrule(lr){4-5} \cmidrule(lr){6-7}
      \cmidrule(lr){8-9} \cmidrule(lr){10-11} \cmidrule(lr){12-13}
      \cmidrule(lr){14-15} \cmidrule(lr){16-17} \cmidrule(lr){18-19}
        & \textbf{4o-m} & \textbf{4o}
        & \textbf{4o-m} & \textbf{4o}
        & \textbf{4o-m} & \textbf{4o}
        & \textbf{4o-m} & \textbf{4o}
        & \textbf{4o-m} & \textbf{4o}
        & \textbf{4o-m} & \textbf{4o}
        & \textbf{4o-m} & \textbf{4o}
        & \textbf{4o-m} & \textbf{4o}
        & \textbf{4o-m} & \textbf{4o} \\
      \midrule
      % \midrule
      % Baseline-Prompting
      Zero-shot     & 22.0  & 32.0  & 10.0  &  9.0  & 17.0  & 25.0  & 10.1  & 19.1  & 12.3  & 24.7  & 11.0  & 21.0  &  4.0  & 18.0  &  6.0  &  7.0  &  5.0  &  6.0  \\
      Few-shot      & 30.0  & 28.0  & 22.0  & 20.0  & 31.0  & 29.0  & \underline{23.6} & 23.6  & \textbf{28.8} & 27.4  & 10.0  &  9.0   & \underline{25.0} & \textbf{24.0} & \textbf{16.0} & 14.0  &  8.0  & 11.0  \\
      CoT~\cite{wei2022chain}           & 21.0  & 39.0  & 13.0  & 10.0  & 26.0  & 30.0  & 18.0  & 28.1  & \textbf{28.8} & 26.0  & \underline{35.0} & 35.0  & 14.0  & \textbf{24.0} &  6.0  & 12.0  & 10.0  & 15.0  \\
      CoT-SC~\cite{wang2022self}      & 20.0  & 37.0  & 11.0  &  6.0  & 20.0  & \textbf{35.0} & 16.9  & 30.3  & \textbf{28.8} & 30.1  & 34.0  & \underline{43.0} & 19.0  & \underline{22.0} & 10.0  & 10.0  & \underline{13.0} & 14.0  \\
      \midrule
      % \midrule
      % Advanced-Prompting
      MultiPersona~\cite{wang2023unleashing}  & 29.0  & 45.0  & 13.0  & 15.0  & 21.0  & 25.0  & 15.7  & 29.2  & 26.0  & \underline{37.0} & \textbf{36.0} & 42.0  & 17.0  & 21.0  &  7.0  & 10.0  & 12.0  & 16.0  \\
      Self-Refine~\cite{madaan2023self}   & \textbf{32.0} & 41.0  & 12.0  & 13.0  & 24.0  & 34.0  & 15.7  & 28.1  & \underline{27.4} & 34.2  & 31.0  & 34.0  & 16.0  & \underline{22.0} &  7.0  & \underline{17.0} & 12.0  & 19.0  \\
      MedPrompt~\cite{chen2024medprompt}     & 29.0  & 34.0  & 14.0  & 11.0  & 30.0  & 26.0  & 13.5  & 22.5  & 20.5  & 26.0  & 34.0  & 22.0  & 18.0  & 16.0  &  6.0  & 14.0  & \underline{13.0} &  9.0  \\
      \midrule
      % \midrule
      % Search-Agent
      SPO~\cite{xiang2025self}           & 19.0  & 31.0  & \underline{25.0} & \underline{31.0} & 20.0  & 30.0  & 22.5  & 29.2  & 19.2  & 32.9  & 32.0  & 36.0  & 14.0  & 19.0  & 11.0  & 15.0  & 11.0  & 16.0  \\
      AFlow~\cite{zhang2024aflow}         & \underline{30.0} & \underline{48.0} & 15.0  & 18.0  & 25.0  & 31.0  & 15.7  & \textbf{34.8} & 24.7  & \textbf{38.4} & 29.0  & 37.0  &  7.0  & 22.0  &  7.0  & 13.0  &  7.0  & \underline{18.0} \\
      \midrule
      % \midrule
      % Multi-Agent
      MedAgents~\cite{tang2023medagents}     & 24.0  & 43.0  & 12.0  & 15.0  & 22.0  & 30.0  & 15.7  & 27.0  & 24.7  & 28.8  &  3.0  &  8.0   & 12.0  & 19.0  &  4.0  &  3.0  &  5.0  &  6.0  \\
      MDAgents~\cite{kim2024mdagents}      & 22.0  & 36.0  & 23.0  & 11.0  & 16.0  & 22.0  & 14.6  & 21.3  & 17.8  & 24.7  &  9.0  &  8.0   & 10.0  & 13.0  &  8.0  &  4.0  &  9.0  &  5.0  \\
      \rowcolor[HTML]{EFEFEF}
      \textbf{Ours} & \textbf{32.0} & \textbf{50.0} & \textbf{35.0} & \textbf{37.0} & \textbf{31.0} & \underline{34.0} & \textbf{25.8} & \underline{31.5} & 26.0  & 28.8  & 32.0  & \textbf{50.0} & \textbf{26.0} & \textbf{24.0} & \underline{14.0} & \textbf{21.0} & \textbf{14.0} & \textbf{19.0} \\
      \bottomrule
    \end{tabular}%
  }
\vspace{-5mm}
\end{table}

\begin{wraptable}{r}{0.5\textwidth}
\vspace{-4mm}
\centering
\caption{Comparison of three medical VQA datasets requiring image-text reasoning. Our method consistently outperforms GPT-4o and multi-agent baselines.}
\label{tab:vqa}
\begin{adjustbox}{width=0.48\textwidth}
% \begin{tabular}{l||ccc}
\begin{tabular}{lccc}
\toprule
\textbf{Method} & \textbf{MedXpert-MM} & \textbf{PMC-VQA} & \textbf{Path-VQA} \\ 
\midrule
% \midrule 
GPT-4o & 24.0\% & 32.0\% & 42.0\% \\ 
MedAgents & 24.0\% & 42.0\% & 48.0\% \\ 
MDAgents & 28.0\% & 54.0\% & 44.0\%\\ 
\textbf{Ours} & \textbf{34.0\%} & \textbf{58.0\%} & \textbf{50.0\%} \\ 
\bottomrule
\end{tabular}
\end{adjustbox}
\vspace{-3mm}
\end{wraptable}

\subsection{Comparison with Multi-Agent, Prompting, and Search-Agent Methods}

We compare our method with three strategies: (i) prior multi-agent LLM frameworks (MedAgents, MDAgents), (ii) prompting-based methods (baseline-prompting and advanced-prompting), and (iii) search-agent systems, across eight challenging medical Q\&A datasets under both GPT-4o-mini and GPT-4o settings.

As shown in Table~\ref{tab:multi-agent}, our method achieves state-of-the-art accuracy on most datasets, outperforming all other methods in 12 of the 18 evaluation columns. We summarize our findings as follows: 
(i) \textbf{Robust gains across all datasets.}  
Our method outperforms prior multi-agent approaches in all 18 comparisons (nine datasets $\times$ two base models), demonstrating strong generalization across diverse tasks and domains.
(ii) \textbf{Superior reasoning under limited model capacity.}  
On GPT-4o-mini, our method surpasses all previous multi-agent methods by a significant margin. For instance, on MMLU-Pro (4o-mini), we achieve 32.0\%, far exceeding MedAgents (3.0\%) and MDAgents (9.0\%), highlighting the effectiveness of our disturbance-enhanced collaboration even with weaker backbones.
(iii) \textbf{Bridging the multi-agent performance gap.}  
Multi-agent frameworks typically struggle on benchmarks such as MMLU-Pro and MedXpert-R, with prior methods (\textit{e.g.}, MDAgent) achieving only 8.0\% on MMLU-Pro (4o) and 4.0\% on MedXpert-R (4o). In contrast, our method achieves 50.0\% and 21.0\% respectively, outperforming all agent-based baselines and matching or exceeding strong prompting and search-based alternatives. 
(iv) \textbf{Limits of multi-agent methods on simpler cases.}  
On the MMLU subset, all multi-agent methods show suboptimal performance due to the simplicity of many test cases.
Most samples in this subset are basic queries that require limited reasoning, reducing the benefits of agent collaboration.

\subsection{Comparison on Medical Visual Question Answer Tasks}

To assess generalization beyond text-based Q\&A, we evaluate our method on medical VQA tasks requiring joint reasoning over clinical images and text. 
Experiments are conducted on three datasets: MedXpert-MM~\cite{zuo2025medxpertqa}, PMC-VQA~\cite{zhang2023pmc}, and PathVQA~\cite{he2020pathvqa}. 
For each, we select 50 samples balanced by complexity: 12 basic, 25 intermediate, and 13 advanced cases, enabling comprehensive evaluation across difficulty levels.

As shown in Table~\ref{tab:vqa}, our method outperforms both the GPT-4o baseline (the base model of our agents) and prior multi-agent frameworks across all benchmarks. 
On MedXpert-MM, it achieves 34\%, exceeding MedAgents and MDAgents by ten and six points, respectively. Similar improvements have been seen on PMC-VQA and Path-VQA. 
These gains demonstrate our method's superior multimodal reasoning, particularly under visual ambiguity.

\subsection{Ablation Study}

To assess the contribution of each component in our framework, we conduct ablation studies on the MedQA dataset using GPT-4o as the base agent model. Following our earlier categorization, we focus on intermediate questions, where silent agreement behavior is most likely to occur, unlike basic cases (single-agent) and advanced ones (forced responses). For each setting, we report: (i) the number of intermediate cases, (ii) overall accuracy, (iii) silent agreement rate, which is defined as the proportion of intermediate cases with no agent response, and (iv) accuracy on non-silent intermediate cases. All silence-related metrics are computed exclusively within the intermediate subset, where such dynamics are more observable.

\vspace{-2.5mm}
\paragraph{Placement of the Catfish Agent.}
We investigate the effectiveness of Catfish Agent placements through four configurations: (i) no Catfish Agent (baseline), (ii) embedded in the Moderator only, (iii) embedded in the Team only, and (iv) embedded in both Moderator and Team (our full configuration). As shown in Table~\ref{tab:ablation-catfish}, introducing the Catfish Agent in either location reduces Silent Agreement and improves accuracy. The best results are achieved when the Catfish Agent is placed in both roles, yielding the highest non-silent accuracy (55.2\%) and the lowest silent rate (17.1\%). These results underscore the complementary value of combining top-down (Moderator) and peer-level (Team) interventions to mitigate silent consensus and foster deeper discussion.

\vspace{-2.5mm}
\paragraph{Tone of the Catfish Agent.} 
We investigate whether the Catfish Agent's tone impacts its effectiveness by comparing a neutral variant with a strategically challenging one, as detailed in Sec.~\ref{sec:intervention-tone}, while keeping the agent embedded in both the Team and Moderator. As shown in Table~\ref{tab:ablation-catfish}, the use of deliberate tone strategies yields higher overall accuracy (50\% vs. 45\%), reduces the Silent Agreement rate (17.1\% vs. 23.3\%), and improves non-silent case accuracy (55.2\% vs. 45.5\%). This highlights tone modulation as a key factor in disrupting premature consensus and encouraging active discussion.

\begin{table}[tp]
\centering
\caption{Ablation study on the Catfish Agent's placement and tone design.}
\label{tab:ablation-catfish}
\begin{adjustbox}{width=0.92\textwidth}
% \begin{tabular}{l||c|c|c|c}
% 
\begin{tabular}{lcccc}
\toprule
\textbf{Configuration} & \textbf{Accuracy $\uparrow$} & \textbf{Intermediate Cases} & \textbf{Silent Rate $\downarrow$} & \textbf{Non-Silent Accuracy $\uparrow$} \\ \midrule
% \midrule
w/o Catfish & 36.0\% & 34 & 61.8\% (21/34) & 38.5\% (5/13) \\
w/ Catfish in Moderator only & 39.0\% & 33 & 51.5\% (17/33) & 37.5\% (6/16) \\
w/ Catfish in Team only & 44.0\% & 30 & 33.3\% (10/30) & 50.0\% (10/20) \\
w/ Catfish (no Tone Design) & 45.0\% & 43 & 23.3\% (10/43) & 45.5\% (15/33) \\
\textbf{w/ Catfish (Full Design)}  & \textbf{50.0\%} & 35 & \textbf{17.1\% (6/35)} & \textbf{55.2\% (16/29)} \\
\bottomrule
\end{tabular}
\end{adjustbox}
\vspace{-5mm}
\end{table}

%\appendix

\section{Case Study}
\label{sec:case_study}

\subsection{Advanced Case}

As illustrated in Figure~\ref{fig:case_advanced}, the diagnostic process begins with multiple specialized teams, each producing structured reports through intra-team collaboration. The Catfish Agent $C^\ast$ monitors interactions and dynamically intervenes when Silent Agreement or conversational bias is detected. It selects an expert role it considers most appropriate (e.g., nephrologist in Figure~\ref{fig:case_advanced}) to raise challenges or provide counterpoints. These interventions are addressed by the team and incorporated into the reasoning trace. The updated report is then passed to the next team for further refinement. After all teams contribute, the Moderator aggregates the insights and, if needed, the Catfish Agent performs a final cross-team critique before the Moderator issues the final decision.

\noindent
% \textbf{Question:} Which of the following management strategies is the most appropriate for this patient? \\
% \textbf{Patient Information:} 43-year-old obese woman with stage 3B CKD, 26-year history of type 1 diabetes (on insulin), and hypertension (on HCTZ 25 mg and lisinopril 40 mg). BP: 140/84 mm Hg. \\
% \textbf{Labs}: Creatinine: 1.7 mg/dL; Potassium: 4.9 mEq/L; ACR: 760 mg/g (↑ proteinuria) \\
\textbf{Question:} A 43-year-old woman with obesity is being assessed for stage 3B chronic kidney disease. She has a 26-year history of type 1 diabetes managed with insulin and hypertension treated with hydrochlorothiazide 25 mg daily and lisinopril 40 mg daily. Her blood pressure is currently 140/84 mm Hg. Laboratory results show a serum creatinine level of 1.7 mg/dL (reference range, 0.6–1.1) and a serum potassium level of 4.9 mEq/L (3.5–5.0). Proteinuria is confirmed with an albumin-to-creatinine ratio of 760 mg/g (<30). Which of the following management strategies is the most appropriate for this patient?
\textbf{Options:} (A) Replace hydrochlorothiazide with dapagliflozin; (B) Add hydralazine to current therapy; (C) Add losartan to current therapy; (D) Increase lisinopril dosage beyond 40 mg daily; (E) Add metoprolol to current therapy; (F) Replace hydrochlorothiazide with canagliflozin; (G) Add amlodipine to current therapy; (H) Replace lisinopril with spironolactone; (I) Replace hydrochlorothiazide with furosemide; (J) Initiate sodium bicarbonate therapy.

\begin{figure}[htb!]
    \centering
    \includegraphics[width=0.99\linewidth]{appendix_figs/fig_advanced.pdf}
    \caption{\emph{Advanced} case example. Interventions from the Catfish Agent leads to a correct decision. Upon detecting premature consensus and inaccurate analysis, the Catfish Agent (as a nephrologist) raises specific concerns, prompting Teams and the Moderator to re-evaluate and ultimately select the correct option.}
    \label{fig:case_advanced}
\end{figure}

\begin{figure}[htb!]
    \centering
    \includegraphics[width=0.99\linewidth]{appendix_figs/fig_intermediate.pdf}
    \caption{\emph{Intermediate} case example illustrating interventions from the Catfish Agent during a multi-round debate. Assigned a fixed domain role, the Catfish Agent monitors team dynamics and raises structured dissent to prevent Silent Agreement, enhancing diagnostic robustness.}
    \label{fig:case_intermediate}
\end{figure}

\subsection{Intermediate Case}

As shown in Figure~\ref{fig:case_intermediate}, a group of expert agents is first recruited, including the Catfish Agent, which is assigned a specific medical role by the Moderator. Each expert independently provides an initial judgment and proposes a preliminary diagnosis. This is followed by multiple rounds of structured debate. During each round, agents sequentially review the initial diagnosis report and are invited to join the discussion by contributing their own perspectives if they disagree or have additional insights. 

Throughout the process, the Catfish Agent monitors for signs of Silent Agreement, overlooked differentials, insufficient justifications, and logical inconsistencies. When such issues arise, the Catfish Agent injects domain-specific challenges calibrated to the group's level of convergence. Targeted agents may revise their responses if they find the critique valid. After each round, a Summary Agent compiles an updated diagnostic report reflecting the latest viewpoints. 

The discussion process terminates once consensus is reached, no substantial divergence follows Catfish interventions, or a predefined round limit is met. Finally, the Moderator makes the ultimate diagnostic decision, optionally consulting the Catfish Agent for additional critique before finalizing the output.

\noindent
\textbf{Question:}  A 24-year-old woman, 8 weeks pregnant, attends her first prenatal visit. She recently immigrated from Africa, has no vaccination records, and works as a babysitter with recent exposure to children with flu and chickenpox. She reports only mild fatigue and nausea. Vitals and physical exam are normal. Which vaccine should she receive now? \\
\textbf{Options:} (A) Tetanus/Diphtheria/Pertussis vaccine; (B) Rabies vaccine; (C) Measles/Mumps/Rubella vaccine; (D) Live-attenuated influenza vaccine; (E) Varicella vaccine; (F) Human papillomavirus vaccine; (G) Intramuscular flu vaccine; (H) Pneumococcal conjugate vaccine; (I) Hepatitis A vaccine; (J) Hepatitis B vaccine.

\subsection{Basic Case}

As shown in Figure~\ref{fig:basic_correct}, this case demonstrates a successful intervention in a basic-complexity question. The Moderator initially provides an incorrect diagnosis $D$, but the Catfish Agent identifies a flaw in the reasoning and offers a concise critique. With this feedback, the Moderator revises the decision and ultimately makes the correct final diagnosis $F$.

\begin{figure}[ht]
    \centering
    \includegraphics[width=0.99\linewidth]{appendix_figs/fig_basic_correct.pdf}
    \caption{A \emph{basic}-level case where the Catfish Agent identifies an oversight in the initial diagnosis and successfully prompts a correction, leading to the correct final decision.}
    \label{fig:basic_correct}
\end{figure}

\subsection{Visual Question Answering Case}

Figure~\ref{fig:case_vqa} presents a successful example where the Catfish Agent injects a targeted perturbation during visual question answering, prompting domain experts to reconsider their initial conclusion. This leads to a course correction and ultimately results in the correct diagnosis.

\begin{figure}[htb!]
    \centering
    \includegraphics[width=0.99\linewidth]{appendix_figs/fig_vqa.pdf}
    \caption{Successful \emph{VQA} case where the Catfish Agent challenges premature consensus by prompting further reflection, guiding the expert team toward the correct diagnosis.}
    \label{fig:case_vqa}
\end{figure}

\subsection{Comparison with DeepSeek-R1}

To further demonstrate the effectiveness of the Catfish Agent, we compare our framework with the strongest baseline model, \texttt{DeepSeek-R1}, using the same clinical question, as shown in Figure~\ref{fig:deepseek_failure} and Figure~\ref{fig:deepseek_vs_ours}. 

In Figure~\ref{fig:deepseek_failure}, \texttt{DeepSeek-R1} conducts a thorough analysis of all available options, yet ultimately fails to select the correct answer. Moreover, it redundantly repeats reasoning patterns across options without meaningful refinement. 

In contrast, Figure~\ref{fig:deepseek_vs_ours} illustrates the final decision stage of our framework's response to the same question. Despite initial incorrect diagnosis among the expert groups, the Catfish Agent identifies a critical flaw and proposes an effective alternative. This intervention successfully prompts the Moderator to revise the initial judgment and reach the correct final decision.

\begin{figure}[ht]
    \centering
    \includegraphics[width=0.99\linewidth]{appendix_figs/fig_deepseek_failure.pdf}
    \caption{\texttt{DeepSeek-R1} failure case. Despite analyzing all answer choices, the model fails to identify the correct one, showing redundancy in reasoning without effective refinement.}
    \label{fig:deepseek_failure}
\end{figure}

\begin{figure}[ht]
    \centering
    \includegraphics[width=0.99\linewidth]{appendix_figs/fig_compare_deepseek.pdf}
    \caption{The final decision stage of our method in response to the same question. Although Group 2 proposes an incorrect option, the Catfish Agent challenges it with an effective alternative, guiding the Moderator toward the correct final decision.}
    \label{fig:deepseek_vs_ours}
\end{figure}

\subsection{Failure Case}

Figure~\ref{fig:case_simple_fail} illustrates a failure case in a basic-complexity question. Despite the Catfish Agent proposing alternative diagnoses and constructively challenging the initial reasoning, the Moderator adheres to the original answer without sufficient reconsideration. This ultimately results in an incorrect final decision, underscoring that the Catfish's interventions, while helpful, can still be overridden in rigid decision-making scenarios.

\begin{figure}[tp]
    \centering
    \includegraphics[width=0.99\linewidth]{appendix_figs/fig_case_basic_failure.pdf}
    \caption{Failure case in a \emph{basic}-level question showing that even with the Catfish Agent's dissent, the Moderator may override critique and finalize an incorrect diagnosis.}
    % Highlights the residual risk of silent anchoring and rigid authority in decision-making pipelines.
    \label{fig:case_simple_fail}
\end{figure}

%\clearpage

\section{Conclusion}
\label{sec:conclusion}

We identify \textbf{Silent Agreement} as a critical failure mode in multi-agent LLM systems for clinical decision making, where agents prematurely converge on diagnoses without sufficient critical analysis. 
To address this, we present the new concept called \textbf{Catfish Agent}, a structured dissent mechanism collaborative reasoning through dynamic, round-based interventions. By these new means, we encourage deeper justification, broader hypothesis exploration, and more robust diagnostics, supported by the proposed \textit{complexity-aware} intervention strategy and \textit{tone-calibrated} intervention mechanism. Experiments on nine public medical Q\&A datasets and three public medical VQA datasets show substantial performance improvements. 
In the future, we plan to investigate efficient coordination strategies that maintain reasoning depth while reducing the inference-time overhead.

{
\small
% \bibliography{reference}

\bibliographystyle{plain}

}

\end{document}